\documentclass[conference]{IEEEtran}
\IEEEoverridecommandlockouts
\usepackage{cite}
\usepackage{amsmath,amssymb,amsfonts}
\usepackage{algorithmic}
\usepackage{graphicx}
\usepackage{textcomp}
\usepackage{xcolor}
\usepackage{url}
\usepackage{hyperref}
\usepackage{booktabs}
\usepackage{multirow}

\usepackage{arydshln}
\usepackage{subcaption}
\usepackage{pifont}

\usepackage{microtype}
\def\BibTeX{{\rm B\kern-.05em{\sc i\kern-.025em b}\kern-.08em
    T\kern-.1667em\lower.7ex\hbox{E}\kern-.125emX}}

\makeatletter
\def\@fnsymbol#1{\@arabic{#1}} 
\makeatother

\makeatletter
\DeclareRobustCommand*{\IEEEauthorrefmark}[1]{%
  \raisebox{0pt}[0pt][0pt]{\textsuperscript{\footnotesize #1}}}
\makeatother

\begin{document}

\title{CogniAlign: Word-Level Multimodal Speech Alignment with Gated Cross-Attention for Alzheimer’s Detection}

\author{
David Ortiz-Perez\IEEEauthorrefmark{1, *},
Manuel Benavent-Lledo\IEEEauthorrefmark{1},
Javier Rodriguez-Juan\IEEEauthorrefmark{1}, \\
Jose Garcia-Rodriguez\IEEEauthorrefmark{1, 2},
David Tomás\IEEEauthorrefmark{3} \\\\
\IEEEauthorblockA{\IEEEauthorrefmark{1}Department of Computer Science and Technology, University of Alicante, Alicante, Spain}
\IEEEauthorblockA{\IEEEauthorrefmark{2}Valencian Graduate School and Research Network of Artificial Intelligence, Valencia, Spain}
\IEEEauthorblockA{\IEEEauthorrefmark{3}Department of Software and Computing Systems, University of Alicante, Alicante, Spain} \\
\IEEEauthorblockA{
\IEEEauthorrefmark{*}Corresponding author -  dortiz@dtic.ua.es \\
Emails: \{dortiz, mbenavent, jrodriguez, jgarcia\}@dtic.ua.es, dtomas@dlsi.ua.es
}
}

\maketitle

\begin{abstract}
Early detection of cognitive disorders such as Alzheimer’s disease is critical for enabling timely clinical intervention and improving patient outcomes. In this work, we introduce CogniAlign, a multimodal architecture for Alzheimer’s detection that integrates audio and textual modalities, two non-intrusive sources of information that offer complementary insights into cognitive health. Unlike prior approaches that fuse modalities at a coarse level, CogniAlign leverages a word-level temporal alignment strategy that synchronizes audio embeddings with corresponding textual tokens based on transcription timestamps. This alignment supports the development of token-level fusion techniques, enabling more precise cross-modal interactions.
To fully exploit this alignment, we propose a Gated Cross-Attention Fusion mechanism, where audio features attend over textual representations, guided by the superior unimodal performance of the text modality. In addition, we incorporate prosodic cues, specifically interword pauses, by inserting pause tokens into the text and generating audio embeddings for silent intervals, further enriching both streams.
We evaluate CogniAlign on the ADReSSo dataset, where it achieves an accuracy of 87.35\% over a Leave-One-Subject-Out setup and of 90.36\% over a 5 fold Cross-Validation, outperforming existing state-of-the-art methods. A detailed ablation study confirms the advantages of our alignment strategy, attention-based fusion, and prosodic modeling. Finally, we perform a corpus analysis to assess the impact of the proposed prosodic features and apply Integrated Gradients to identify the most influential input segments used by the model in predicting cognitive health outcomes.
\end{abstract}



\section{Introduction}

Cognitive decline is a well-documented aspect of the aging process~\cite{deary2009age}. As individuals age, abilities such as reasoning, memory, and communication typically show signs of gradual deterioration. While this decline is often a normal consequence of aging, it can become more severe in the presence of specific neurological conditions. Among these, cognitive disorders like dementia~\cite{duong2017dementia,celsis2000age} are particularly prominent, with Alzheimer’s disease being the most prevalent and widely studied form. Early detection of such conditions is essential, as timely medical intervention can help slow cognitive deterioration and improve the patients’ quality of life. This is especially relevant in the context of an aging global population~\cite{di2018towards}, where the rising proportion of older adults is expected to lead to a significant increase in dementia cases~\cite{ageing}. These demographic trends highlight the need for early, accessible, and accurate diagnostic methods.

In this context, deep learning techniques have emerged as powerful tools for automating early detection of cognitive decline~\cite{venugopalan2021multimodal,liu2014early}. These approaches have shown considerable promise across a wide range of healthcare applications, including the summarization of clinical interviews~\cite{cabello2024automated}, pain level estimation~\cite{salekin2021multimodal,benavent2023comprehensive}, or brain tumor detection~\cite{chattopadhyay2022mri}. Automating such tasks offers several key advantages. Most notably, it facilitates continuous monitoring and assessment of a patient’s behavior, enabling proactive identification of cognitive changes without the need for scheduled clinical visits. This capability is particularly valuable in healthcare environments, where professionals often contend with demanding workloads and increased risk of burnout~\cite{lin2021long}. Furthermore, deep learning enables a more personalized approach to diagnosis by modeling the individual progression of each patient. These models are capable of continuously analyzing temporal patterns in cognitive behavior, providing detailed insights into the trajectory of decline over time~\cite{gatla2024innovative,esteva2022prostate}. Such progression-aware analysis allows for the early identification of accelerated or atypical deterioration, enabling timely alerts to healthcare professionals and facilitating prompt clinical evaluation when necessary.

Recent advances in deep learning have increasingly emphasized the integration of multiple sources of information rather than relying solely on a single input modality. These sources, commonly referred to as modalities, can include audio, text, video, or various forms of medical data (MRI, ECG, EHR, etc). Multimodal learning, which combines different modalities, enables a more comprehensive and nuanced understanding of complex tasks by providing complementary and context-rich perspectives on the same underlying task~\cite{xu2023multimodal,han2023survey}. Each modality offers unique and complementary information that enhances the overall interpretability of a task. For instance, in speech analysis, the audio modality captures how individuals articulate their thoughts, providing access to prosodic cues such as intonation, hesitations, and emotional tone. Meanwhile, its textual transcription adds further insight by capturing lexical selection, syntactic structure, and semantic content. This multimodal combination is particularly effective in tasks such as irony or sarcasm detection, where both verbal and non-verbal cues, i.e. what is said and how it is said, are critical for accurate understanding~\cite{boutsikaris2024comparative,tomas2023transformer}.

Beyond speech, the value of multimodal approaches extends broadly across healthcare applications. While medical imaging can provide crucial diagnostic information, it often requires invasive or resource-intensive procedures~\cite{zhou2021review,liu2021review}. In contrast, audio and textual data offer a non-intrusive and low-cost alternative, enabling continuous patient monitoring without disrupting daily activities. Moreover, multimodal systems enhance both performance and robustness by integrating complementary information. By reducing reliance on a single modality, these systems remain effective even when one source is noisy, incomplete, or missing, drawing on other modalities to support reliable and accurate decision-making~\cite{liu2021comparing,ma2022multimodal}.

In response to these needs, we introduce CogniAlign, a multimodal model for Alzheimer’s disease detection that leverages audio and text, two complementary and non-intrusive modalities, to capture indicative patterns of cognitive decline. CogniAlign yields an accuracy of 87.35\% over a Leave-One-Subject-Out (LOSO) and of 90.36\% over a 5 fold Cross-Validation (CV) on the ADReSSo dataset~\cite{luz21_interspeech}, outperforming existing state-of-the-art methods. Unlike prior work that often treats modalities independently and fuses them only at later stages, our approach performs a fine-grained word-level alignment between audio and text embeddings, allowing for more coherent and semantically grounded fusion. To enable this, we use the Whisper model~\cite{radford2022whisper} to transcribe the audio recordings and extract word-level timestamps. These timestamps guide the temporal alignment between textual tokens and audio embeddings.

Additionally, we enrich the model with prosodic information by identifying speech pauses and inserting corresponding pause tokens into the transcription, which are then aligned with audio features to enhance the multimodal representation and boost both unimodal and multimodal performance. To enable effective multimodal fusion, we propose a gated cross-attention fusion strategy combined with a word-level alignment mechanism, ensuring that tokens from each modality correspond to the same speech segments. This approach significantly enhances the model's ability to capture cross-modal interactions and improves overall performance. Finally, we conducted a corpus analysis to examine the influence of the proposed prosodic features within the dataset, and we analyzed the Integrated Gradients to identify the most relevant segments of each sample that the model leverages when predicting Alzheimer's disease and cognitive health. The implementation code is publicly available on GitHub~\footnote{\url{https://github.com/davidorp/CogniAlign}}.

In summary, our main contributions are as follows:

\begin{itemize}
    \item We propose CogniAlign, a multimodal model for Alzheimer’s disease detection that effectively integrates audio and text modalities using a gated cross-attention fusion strategy, achieving state-of-the-art performance.
    \item We introduce a word-level alignment strategy that synchronizes audio and text embeddings, enabling a more semantically grounded multimodal fusion.
    \item We incorporate prosodic features, specifically speech pauses, to enrich the representation and improve both unimodal and multimodal performance.
    \item We conduct a comprehensive ablation study to evaluate the impact of each modality, fusion strategy, and architectural design choice.
    \item We perform a corpus analysis to examine the influence of the proposed prosodic features within the dataset. Additionally, we analyzed the Integrated Gradients to identify the most relevant segments of each sample that the model considers when predicting Alzheimer's disease and cognitive health.
\end{itemize}

 The remainder of this paper is organized as follows: Section~\ref{sec:related} reviews the relevant literature on the topic; Section~\ref{sec:method} describes the methodology employed throughout this work; Section~\ref{sec:exp} presents the experimental setup and results; Section~\ref{sec:explainable} introduces an analysis of the corpus and different explainability methods; Section~\ref{sec:limitations} discusses the main limitations regarding data availability of this work; finally, Section~\ref{sec:conclusions} outlines the conclusions drawn from this study.
\section{Related Works}\label{sec:related}

This section reviews the current literature on cognitive decline detection, focusing on deep learning approaches and multimodal fusion strategies.

\subsection{Cognitive Decline Detection using Deep Learning Techniques}

Deep learning techniques have been extensively applied to cognitive decline estimation. While some approaches rely on intrusive modalities such as medical imaging~\cite{liu2021review,zhou2021review}, non-intrusive data sources, particularly speech and language, are gaining traction due to their ease of collection and potential diagnostic value~\cite{qi2023noninvasive,yang2022deep}. Although deep learning shows great potential, these models typically require large volumes of data to achieve robust performance. However, in the clinical domain, data availability remains a significant challenge due to privacy concerns, and the sensitive nature of medical information.

In the context of cognitive decline, several datasets have been developed, each targeting different diseases, modalities, and languages. For example, the DementiaBank Pitt Corpus~\cite{lanzi2023dementiabank} provides a rich collection of speech recordings and corresponding transcriptions from individuals with dementia, and cognitively healthy controls. These recordings span multiple tasks, including picture description, verbal fluency, and recall exercises. In contrast, the ADReSSo dataset~\cite{luz21_interspeech} offers audio-only recordings from individuals diagnosed with Alzheimer’s disease and healthy controls without accompanying transcriptions. Expanding to multilingual contexts, the Taukadial dataset~\cite{luz24_interspeech} includes audio recordings from Chinese and English-speaking participants, encompassing individuals with Mild Cognitive Impairment (MCI) and healthy subjects.

One common limitation across these datasets is their relatively small size, with the largest containing fewer than 500 samples (DementiaBank Pitt Corpus). The scarcity of labeled data poses a significant barrier to training large-scale deep learning models, underscoring the importance of data-efficient architectures and transfer learning strategies in this domain.

Despite this limitation, recent studies have increasingly leveraged deep learning techniques for cognitive decline detection. Many of these works consistently report superior performance when using textual data over audio~\cite{app13074244,10389785,bang2024alzheimer}, suggesting that the semantic structure of language, i.e. how individuals organize and express their thoughts, may be more indicative of the cognitive status than prosodic or acoustic features alone. Additionally, some of these studies also demonstrate that combining both modalities yields improved results compared to unimodal approaches, underscoring the complementary nature of audio and textual information~\cite{bang2024alzheimer,ying2023multimodal,rohanian21_interspeech,10389785}.

Among multimodal approaches, joint fusion strategies are the most widely adopted~\cite{ortizperez2024deepinsightscognitivedecline}. These frameworks typically involve training separate unimodal encoders for each modality, followed by a fusion step, usually consisting of a simple feature concatenation. In the case of the textual modality, pretrained language models such as BERT~\cite{devlin2019bert}, DistilBERT~\cite{sanh2019distilbert}, and RoBERTa~\cite{DBLP:journals/corr/abs-1907-11692} are frequently employed. In audio, Wav2Vec2~\cite{baevski2020wav2vec}, Audio Spectrogram Transformers~\cite{gong21b_interspeech,gong_psla}, Mel Spectrograms, or handcrafted acoustic features like eGeMAPS~\cite{7160715} are commonly employed. However, a key limitation of many of these approaches is the absence of explicit cross-modal interactions, as each modality is processed independently prior to fusion.

For instance, in~\cite{bang2024alzheimer}, the authors utilize Wav2Vec2 for audio feature extraction, and BERT for text while incorporating semantic knowledge from large language models, specifically ChatGPT~\footnote{\url{https://chatgpt.com/}}. The fusion strategy involves joint fusion via concatenation.

Similarly in \cite{altinok2024explainable}, the authors use Mel Spectrograms processed with a Vision Transformer (ViT) model~\cite{dosovitskiy2020image} to represent the audio modality. Meanwhile, textual embeddings are obtained using the BERT model. Fusion is achieved through an attention mechanism that integrates both modalities, which outperforms the extended concatenation-based fusion.

\subsection{Multimodal Fusion Strategies}

Recent advancements in deep learning have also embraced multimodal learning paradigms, reflected in the rising development of foundational models such as GPT 4, LLaMA~\cite{grattafiori2024llama3}, or LLaVA~\cite{liu2023visual}. This trend reflects a broader shift toward integrating multiple modalities to enrich data representations, where each modality contributes complementary perspectives. By leveraging cross-modal correlations, multimodal systems enable a more comprehensive understanding of complex tasks and generally lead to improved performance~\cite{10834497,li2024multimodal}.

The incorporation of multiple modalities can be achieved through a variety of methodologies. One approach involves training large end-to-end multimodal architectures, such as VideoBERT~\cite{sun2019videobert} or VideoCLIP~\cite{xu2021videoclip}. However, a more prevalent approach involves constructing models by integrating independently trained unimodal encoders~\cite{bang2024alzheimer,altinok2024explainable,app13074244,10389785}, followed by the application of dedicated fusion mechanisms to combine their outputs.

Fusion strategies are categorized into different groups including early, late, or joint among others~\cite{gao2020survey,GANDHI2023424}. Early fusion involves merging modalities before individual processing, which requires a shared latent space, a constraint that may not be feasible in many scenarios. Other fusion strategies allow modalities to be processed independently before integration, often through diverse mechanisms such as concatenation, arithmetic operations, or attention. Although these approaches provide flexibility and modularity, they generally overlook the temporal and semantic alignment between modalities.

Several recent studies have attempted to improve multimodal integration through alignment strategies that synchronize audio and textual streams. In~\cite{gu-etal-2018-multimodal}, audio is segmented into word-level Mel spectrograms and paired with corresponding textual tokens, with fusion performed via a GRU. Similarly,~\cite{rohanian2019detecting} aligns text, audio, and visual modalities at the word level for depression detection, combining them using an LSTM. Although both works ensure temporal coherence, their reliance on sequential models limits the flexibility to model complex inter-modal relationships.

The WISE model~\cite{shen2020wise} adopts a Transformer-based interaction module to process word-aligned audio and textual sequences. Fusion is implemented via concatenation followed by self-attention, providing richer interdependencies than purely sequential models. However, it lacks directional cross-attention mechanisms that enable explicit querying between modalities, potentially restricting the model's ability to emphasize complementary information.

Other approaches, such as~\cite{xue2024fusing}, focus on sentence-level alignment for depression detection. In this work, embeddings from BERT and corresponding audio segments are combined via concatenation before classification. While this offers semantic alignment at a broader level, it sacrifices the granularity and token-level precision that fine-tuned fusion methods may provide.

While these studies demonstrate the utility of temporal alignment between modalities, they consistently rely on simple fusion techniques such as concatenation or sequential modeling. Few exploit attention-based mechanisms, and none incorporate directional cross-attention, which enables modality-specific querying. As a result, the potential of fine-grained alignment is underexplored, leaving a gap that CogniAlign addresses by combining word-level synchronization with expressive attention-driven fusion.
\section{CogniAlign Architecture}\label{sec:method}

\begin{figure*}[htpb]
    \centering
    \includegraphics[width=\textwidth]{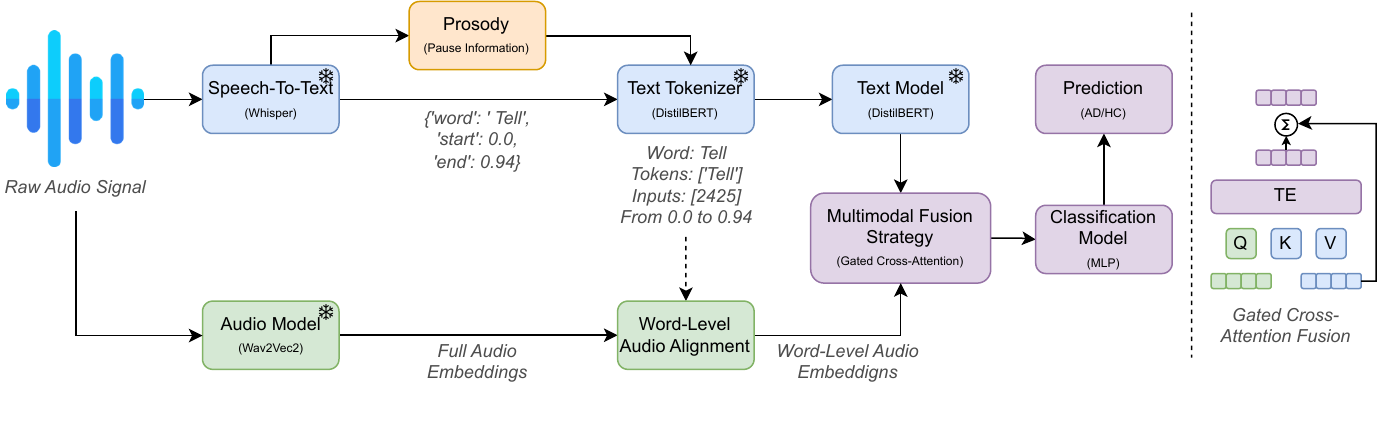}
\caption{\textbf{Overview of the CogniAlign architecture.} Audio recordings from the ADReSSo dataset are transcribed with Whisper, extracting word-level timestamps and prosodic cues (pauses). This enables temporal alignment between textual and audio embeddings at the word level. Aligned features are fused through a Gated Cross-Attention Transformer Encoder (\texttt{TE}), shown on the right of the figure, where textual embeddings serve as queries (\texttt{Q}) and audio embeddings as keys/values (\texttt{K/V}). A learnable gating mechanism regulates the integration of attended features, as illustrated in the right panel. An MLP processes fused representations for Alzheimer's detection. Green blocks represent text components, blue blocks represent audio components, and purple blocks denote multimodal fusion. Frozen pre-trained models are indicated with a snowflake symbol. Better viewed in color.}
    \label{fig:overview}
\end{figure*}

This section presents the CogniAlign architecture proposed in this study. The framework is composed of four main components: a textual encoder, an audio encoder with temporal alignment, a prosody information extractor, and a multimodal fusion strategy. An overview of the complete pipeline is presented in Figure~\ref{fig:overview}. Audio features are temporally aligned to textual tokens using word-level timestamps, enabling synchronized multimodal fusion.

\sloppy
To support fine-grained multimodal modeling, CogniAlign is built upon the Transformer architecture~\cite{10.5555/3295222.3295349}. Transformers are particularly well-suited for capturing contextual and cross-modal relationships, allowing each token to attend selectively to others. This expressiveness makes them ideal for tasks involving subtle cues, such as those in clinical speech data~\cite{NERELLA2024102900,info:doi/10.2196/59505}. Beyond their architectural flexibility, Transformers have consistently outperformed recurrent and convolutional models in a wide range of natural language and multimodal benchmarks~\cite{sarmadi2024comparative,khan2022transformers, ortizperez2024deepinsightscognitivedecline}.

\subsection{Textual Encoder}\label{subsec:textual}
CogniAlign employs a frozen text encoder to extract text features. Freezing the language model has two main advantages: (1) it leverages powerful semantic representations from large-scale pretraining, and (2) prevents overfitting in clinical domains where annotated data is limited~\cite{du2024enhancing,botelho24_interspeech}.

Specifically, textual features are extracted using the frozen DistilBERT model, selected for its performance and computational efficiency, as demonstrated in Section~\ref{subsec:ablation}. The model outputs contextual embeddings for each subword token while preserving the sequence structure from the word-aligned transcription.

\subsection{Audio Encoder with Temporal Alignment}

The audio encoder in CogniAlign employs a frozen Wav2Vec2 model to extract contextualized acoustic representations from the input signal.

To enable fine-grained multimodal fusion, CogniAlign performs a temporal alignment between the audio and textual modalities at the word level. This alignment addresses the mismatch in granularity between the two streams: while Wav2Vec2 produces dense frame-level embeddings every 20 milliseconds, textual models like BERT generate much sparser token-level representations.

To address this, we propose a temporal alignment strategy based on word-level timestamps extracted using Whisper. The transcription \( T \) is represented as a sequence of word-timestamp pairs:
\[
T = \{ (w_1, [t_{1}^{\text{start}}, t_{1}^{\text{end}}]), 
\ldots, (w_N, [t_{N}^{\text{start}}, t_{N}^{\text{end}}]) \},
\]
where \( w_i \) denotes the \( i \)-th word in the transcription, and \( t_i^{\text{start}} \) and \( t_i^{\text{end}} \) represent the start and end times (in seconds) during which the word is spoken in the audio signal.

Given frame-level audio embeddings \( f_j \) produced by Wav2Vec2, each associated with timestamp \( t_j \), we construct word-level audio embeddings by averaging all frames within the corresponding word's interval:
\[
\mathbf{a}_i^{\text{word}} = \text{MeanPooling}(\{ f_j \mid t_j \in [t_i^{\text{start}}, t_i^{\text{end}}] \}).
\]

This results in a set of word-aligned audio embeddings \( \mathbf{A}^{\text{word}} = \{ \mathbf{a}_1^{\text{word}}, \ldots, \mathbf{a}_N^{\text{word}} \} \). For consistency in alignment, when a word is split into multiple subword tokens by the language model tokenizer, the same audio embedding is assigned to all corresponding tokens.

The resulting audio embeddings are temporally and semantically aligned with the corresponding textual tokens. This alignment enables fine-grained fusion through attention mechanisms and facilitates more effective modeling of cross-modal dependencies.

\subsection{Prosodic Information}

\begin{figure}
    \centering
    \includegraphics[width=0.6\linewidth]{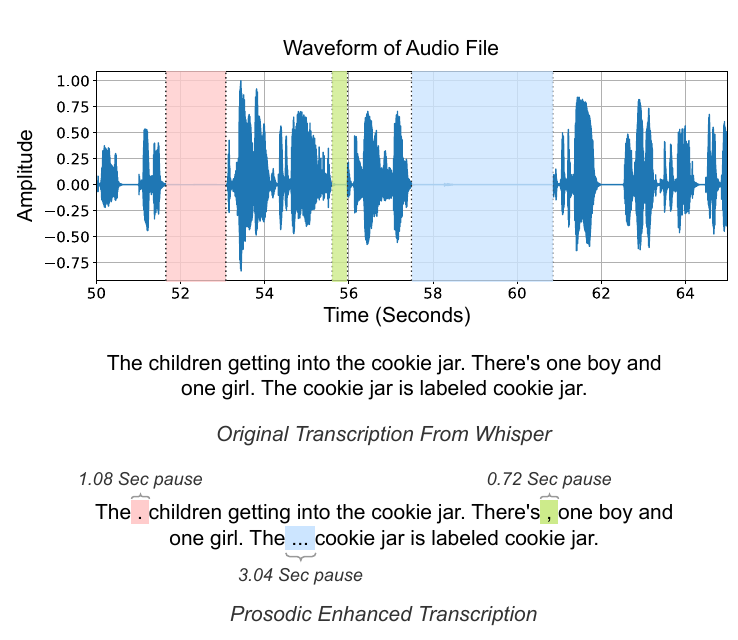}
    \caption{\textbf{Prosodic augmentation pipeline.} Pauses, detected using Whisper word-level timestamps (and also visible as silent regions in the waveform), are inserted into the transcription as punctuation marks (comma, period, ellipsis) based on duration. Inserted pauses, shown in green, enhance the original transcription with prosodic cues.}
    \label{fig:prosodic}
\end{figure}

To capture rhythm and hesitations from natural speech, prosodic information, particularly speech pauses, is extracted and incorporated into the model. These pauses are detected using the word-level timestamps provided by the Whisper model by measuring the temporal gap between consecutive words in the transcription. This process is visually summarized in Figure~\ref{fig:prosodic}.

Pauses are categorized based on their duration: between 0.5 and 1 second are marked with a comma (\emph{,}), those between 1 and 1.5 seconds with a period (\emph{.}), and those longer than 1.5 seconds with an ellipsis (\emph{...}). These symbols are inserted directly into the textual transcription, expanding the sequence passed to the tokenizer. As a result, the text encoder processes additional pause tokens that mirror the prosodic structure of the original speech, enabling the model to capture not only lexical content but also hesitation patterns and rhythm, which are relevant for cognitive decline detection~\cite{yuan20_interspeech,zhu21e_interspeech,rohanian21_interspeech}.

In parallel, each detected pause is assigned a corresponding audio embedding. For each pause interval $[s_p, e_p]$, the embedding is computed by averaging the frame-level audio features $f_j$ that fall within the pause segment:
\[
a_{\text{pause}} = \frac{1}{N_p} \sum_{j: t_j \in [s_p, e_p]} f_j,
\]
where $N_p$ denotes the number of frames associated with the silence.

Audio pause embeddings are inserted into the audio sequence at the corresponding positions, ensuring that pause tokens in the text are temporally aligned to audio representations. Despite the increase number of tokens, the consistent treatment across modalities preserves alignment and enables fine-grained attention-based fusion over lexical content and prosodic cues.

\subsection{Gated Cross-Attention Fusion Strategy}

After processing each modality independently and aligning them at the word level, CogniAlign integrates the audio and textual streams using a Gated Cross-Attention Fusion mechanism. This fusion strategy enables token-level cross-modal interaction through a single Transformer Encoder layer.

Let $\mathbf{A}, \mathbf{T} \in \mathbb{R}^{L \times d}$ denote the audio and textual token sequences, respectively, where $L$ is the number of tokens and $d = 768$ is the embedding dimensionality, consistent with the output sizes of DistilBERT and Wav2Vec2. Based on results from our ablation study (Section~\ref{subsec:ablation}), we designate the audio embeddings $\mathbf{A}$ as the \textit{queries} and the textual embeddings $\mathbf{T}$ as the \textit{keys} and \textit{values} in the Transformer cross-attention mechanism:
\begin{equation}
    \mathbf{H}_{\text{att}} = \text{Attention}(\mathbf{A}, \mathbf{T}, \mathbf{T}).
\end{equation}

This configuration is both empirically and theoretically motivated. In Transformer-based cross-attention, the key/value modality provides the main source of contextual information, while the query determines where to focus~\cite{khan2025memocmt,liu2023tacfn}. Given the superior performance of textual features in unimodal settings, maintaining them as the key/value modality allows their semantic richness to guide the fusion. Meanwhile, the audio stream provides complementary prosodic and paralinguistic cues.

Following the attention step, a learnable gating mechanism is applied to modulate the integration of the attended and original audio features~\cite{xu2023multimodal}. This is achieved by implementing an element-wise sigmoid gate:
\begin{equation}
    \mathbf{G} = \sigma(\mathbf{W}_g \mathbf{H}_{\text{att}} + \mathbf{b}_g),
\end{equation}
Where $\mathbf{W}_g$ and $\mathbf{b}_g$ are learnable parameters, and $\sigma(\cdot)$ denotes the sigmoid activation.

The final fused representation is obtained via a gated residual connection:
\begin{equation}
    \mathbf{H} = \mathbf{G} \odot \mathbf{H}_{\text{att}} + (1 - \mathbf{G}) \odot \mathbf{A},
\end{equation}
Where $\odot$ denotes the element-wise (Hadamard) product.

The resulting multimodal embeddings $\mathbf{H}$ capture both the contextualized textual content and the prosodic structure of the input speech. These embeddings are passed directly to a lightweight MLP classifier to perform Alzheimer's disease detection.

\section{Experiments}\label{sec:exp}

This section presents the experimental framework adopted in this study and the results to assess the performance of the proposed models. In addition to the main experiments, an ablation study is conducted to examine the contribution of individual modalities and fusion strategies. The results are benchmarked against state-of-the-art methods using the ADReSSo dataset.

\subsection{Experimental Setup}

\noindent\textbf{Dataset:} The experiments in this work are conducted using the ADReSSo dataset~\cite{luz21_interspeech}, specifically designed for Alzheimer’s disease detection. It consists of audio recordings of participants describing the \emph{Cookie Theft Picture} from the Boston Diagnostic Aphasia Examination~\cite{becker1994natural,goodglass2001bdae}. The dataset includes a total of 237 samples, with 118 from individuals diagnosed with Alzheimer’s disease and 119 from cognitively healthy controls. 

\noindent\textbf{Implementation Details:} The CogniAlign architecture is implemented using a single Transformer Encoder layer with 768 hidden units, 12 attention heads, and an intermediate size of 3072. This configuration aligns with the original architecture of DistilBERT and Wav2Vec2, ensuring compatibility across modalities. All unimodal models, whether textual or audio-based, are equipped with a single Transformer Encoder layer and employ a mean pooling strategy to generate fixed-length representations, maintaining consistency across the architecture.

To evaluate the performance of unimodal textual models individually, we used the best configurations preceding the classification layers. BERT, RoBERTa, and DistilBERT share the same transformer dimensions as CogniAlign (768 hidden size, 12 attention heads, and 3072 intermediate size). In contrast, larger models maintain their native configurations: Mistral~\cite{jiang2023mistral7b} (4096, 32, 14,336), Qwen~\cite{qwen2,qwen2.5} (3584, 28, 18,944), and Stella~\cite{zhang2025jasperstelladistillationsota} (1536, 28, 8960).  

On the audio side, the Wav2Vec2 encoder mirrors the configuration of the BERT-like models. The Transformer encoders for Mel spectrograms and eGeMAPS features adopt the same 768-dimensional size for consistency across audio representations. To project Mel spectrograms (80 channels) and eGeMAPS (88 features) into the shared feature space, a residual CNN upscaling module is applied prior to the Transformer. This module consists of three 2D convolutional blocks with batch normalization, ReLU activations, and dropout and is trained jointly with the rest of the model.

\noindent \textbf{Training Details:} CogniAlign has been implemented in PyTorch, and all experiments were executed on NVIDIA RTX 4090 GPUs. Models were trained using the AdamW optimizer with a weight decay of 0.1 and an initial learning rate of $2 \times 10^{-5}$.  A warmup phase was applied over the first 20 epochs, followed by a cosine learning rate schedule.

\noindent \textbf{Evaluation Details:} To identify the most effective combination of modalities and hyperparameters, we conducted an ablation study using 5-fold cross-validation. This training setup ran for a maximum of 200 epochs with a batch size of 32, and early stopping was applied with patience of 15 epochs to mitigate overfitting.

The hyperparameter configuration selected in this ablation stage has been unchanged in every subsequent experiment, covering the regression task and the additional evaluation on the ADReSS dataset.

For comparison with state‑of‑the‑art methods, we report metrics under both 5‑fold Cross‑Validation (CV) and LOSO setups. The LOSO configuration employed the same training parameters as the 5‑CV setup but used a fixed 60‑epoch schedule without early stopping, as no validation set was available. Notably, most prior studies use varying cross‑validation strategies, ranging from 70/30 train–test splits to different k‑fold configurations, which can lead to inconsistencies in model evaluation~\cite{ying2023multimodal}.

To address these discrepancies, we follow the baseline approach~\cite{luz21_interspeech}, adopting LOSO as our primary protocol for comparison because it guarantees replicability and ensures a more robust, fair evaluation. Beyond mitigating validation inconsistencies, LOSO performs inference on each subject independently, ensuring that every test sample is entirely isolated from the training data. Unlike traditional k‑fold cross‑validation, LOSO avoids arbitrary data splits and eliminates subject overlap between training and testing sets, providing a more realistic assessment of a model’s generalisability to previously unseen individuals. Meanwhile, the 5‑CV protocol is retained for the ablation study owing to its lighter computational cost, which allows rapid, exhaustive exploration of modality combinations and hyper‑parameter settings.

\subsection{Comparison with State-Of-The-Art Methods}

To assess the effectiveness of the proposed methodology, we compare its performance against current state-of-the-art approaches. Table~\ref{tab:sota} reports accuracy, F1-Score, precision and recall results on the ADReSSo dataset, covering audio-based, text-based, and multimodal models. For consistency with prior works, we include results for unimodal settings alongside multimodal approaches, thereby providing a comprehensive view of our model's relative performance. The unimodal models included correspond to those that achieved the highest performance in our ablation study, as discussed in Section~\ref{subsec:ablation}.

\begin{table*}[htpb]
\centering
\resizebox{\linewidth}{!}{%
\begin{tabular}{lccccccccccccc}
\toprule
\multirow{2}{*}{Work} & & \multicolumn{4}{c}{Audio} & \multicolumn{4}{c}{Text} & \multicolumn{4}{c}{Multimodal} \\
\cmidrule(lr){3-6} \cmidrule(lr){7-10} \cmidrule(lr){11-14}
& Protocol & Acc & F1 & Pr & Rec & Acc & F1 & Pr & Rec & Acc & F1 & Pr & Rec \\\midrule
Wang et al., 2021~\cite{wang21ca_interspeech} & 10CV & 75.30 & 76.00 & 77.40 & 74.70 & 73.50 & 73.50 & 75.30 & 73.60 & 77.20 & 77.60 & 78.70 & 74.00 \\
Rohanian et al., 2021~\cite{rohanian21_interspeech} & - & 68.00 & - & - & - & 81.00 & - & - & - & 84.00 & - & - & - \\
Cui et al., 2023~\cite{10389785} & 80/20 & - & 76.30 & - & - & - & 82.60 & - & - & - & 89.40 & - & - \\
Ying et al., 2023~\cite{ying2023multimodal} & 10CV & 71.20 & 73.10 &  73.10 & 71.00 & 78.90 & 79.00 &  83.10 & 79.00 & 83.70 & 83.60 &  85.30 & 84.00 \\
Priyad. et al., 2023~\cite{app13074244} & 70/30 & 78.90 & - & - & - & {88.70} & - & - & - & - & - & - & - \\
\textbf{CogniAlign (Ours)} & 5CV & {80.12} & {79.46} & {79.88} & {81.40} & 86.77 & {86.59} & {87.07} & {86.83} & {90.36} & {90.11} & 90.15 & {90.76} \\
\midrule
Baseline, 2021~\cite{luz21_interspeech} & LOSO & \textbf{78.92} & - & - & - & 80.72 & - & - & - & - & - & - & - \\
Bang et al., 2024~\cite{bang2024alzheimer} & LOSO & 69.01 & 70.39 & 70.60 & 70.42 & \textbf{83.10} & \textbf{83.10} & 83.10 &\textbf{ 83.10} & 87.32 & 87.25 & 88.06 &\textbf{ 87.32} \\
\textbf{CogniAlign (Ours)} & LOSO & 74.70 & \textbf{76.40} & \textbf{74.73}&\textbf{78.16}& 82.53 & 82.63 &\textbf{86.25} & 79.31 & \textbf{87.35} & \textbf{87.57} & \textbf{90.24} & 85.06 \\
\bottomrule
\end{tabular}
}
\caption{Comparison of state-of-the-art models on the ADReSSo dataset across audio, text, and multimodal modalities. The top half presents results under various evaluation protocols, including our 5CV setting. The bottom half compares models using the main evaluation metric (LOSO), which enables a fair comparison. Bold values indicate the best performance under LOSO. Dashes indicate unreported evaluation protocols or results. For simplicity, ``Acc'' denotes accuracy, ``Pr'' denotes precision, and ``Rec'' denotes recall.}

\label{tab:sota}
\end{table*}

The results demonstrate the effectiveness of the CogniAlign method, which outperforms existing state-of-the-art approaches. Under the LOSO setup, it achieves an accuracy of 87.35\% and an F1 score of 87.57\% using the multimodal approach. In the 5-fold CV setup, it attains an accuracy of 90.36\% and an F1 score of 90.11\%. In this setup, where ablation studies were conducted to optimize unimodal configurations, our CogniAlign unimodal models outperform almost all existing state-of-the-art approaches. More importantly, the multimodal fusion strategy achieves the best performance across both evaluation setups (LOSO and 5-fold CV), even without relying on the strongest unimodal results in the LOSO setup. Regarding the main comparison setup, LOSO, CogniAlign achieves the best multimodal results in all metrics but recall. We attribute the slight drop in recall to the model's conservative prediction strategy, which prioritizes precision over capturing all positive cases.     This highlights the effectiveness of our gated cross-attention mechanism with word-level fusion and underscores the value of integrating multiple modalities in a coordinated manner, rather than treating them as independent sources of information.

In contrast, several existing works adopt less integrated fusion strategies. For instance, the approach presented in~\cite{bang2024alzheimer}, as well as those in~\cite{ying2023multimodal} and~\cite{10389785}, rely on simple concatenation of features processed by separate models, each handling input modalities independently. A similar technique is employed in~\cite{rohanian21_interspeech}, albeit incorporating a gating mechanism to regulate the information flow. Meanwhile, the study in~\cite{wang21ca_interspeech} explores attention-based methods. However, it still treats input data from each modality in isolation. Furthermore, the baseline~\cite{luz21_interspeech} and the method proposed in~\cite{app13074244} do not incorporate any form of multimodal fusion. These comparisons underscore the limitations of approaches that fail to fully integrate multimodal information and further validate the strength of our proposed fusion strategy in capturing cross-modal relationships effectively.

Regarding the textual modality, our selected model is DistilBERT P, which integrates pause token information into the DistilBERT architecture. This implementation is detailed in subsection~\ref{subsec:ablation}. This model exhibits strong performance, achieving an accuracy of 82.53\% and an F1 score of 82.63\% under the LOSO setup, and an accuracy of 86.77\% with an F1 score of 86.59\% in the 5 CV. Although competitive, this model does not achieve the top accuracy overall. A related study presented in~\cite{app13074244}, which utilizes the RoBERTa model, reports an accuracy of up to 88.7\%. Despite its strong performance in this study, where it was fine-tuned end-to-end, RoBERTa underperformed in our experiments, where it was kept frozen for consistency. This methodological difference likely accounts for the variance in performance outcomes.

For the audio modality, our selected model is Wav2Vec2 P, which utilizes Wav2Vec2 embeddings aligned at the word level and augments them with pause token representations derived from timestamp analysis. The implementation of this approach is detailed in subsection~\ref{subsec:ablation}. This model achieves an accuracy of 76.40\% and an F1 score of 76.40\% under the LOSO setup, achieving the state-of-the-art performance on F1, precission and recall. In the 5 fold CV, this model achieves an accuracy of 80.12\% with an F1 score of 79.46\%, outperforming the rest of the proposed methodologies. Its performance highlights the benefit of synchronizing audio features with semantic boundaries rather than processing audio as a continuous stream.

Although Wav2Vec2 P is technically a unimodal model since it relies solely on audio embeddings for classification, it indirectly benefits from transcriptions to obtain precise word-level timestamps via Whisper. These timestamps enable the construction of semantically meaningful audio tokens aligned with spoken words and pause intervals. This preprocessing step is used solely for temporal alignment and does not introduce textual features into the model itself, preserving the unimodal nature of the classifier.

Compared to conventional models that operate over fixed-size audio windows without alignment, this word-level segmentation approach facilitates the capture of meaningful acoustic patterns associated with cognitive decline. Nevertheless, consistent with observations in prior work, audio-only models still trail text-based approaches in overall performance, suggesting that lexical-semantic content offers richer cues for Alzheimer's disease detection.

\subsection{Regression Task}

In addition to a binary label, indicating either cognitive health or Alzheimer's disease, the ADReSSo dataset also provides a numerical score ranging from 0 to 30. This score corresponds to the result of the Mini-Mental State Examination (MMSE), a widely used clinical tool in the healthcare field for assessing cognitive impairment~\cite{arevalo2015mini}. The MMSE consists of a series of eleven questions designed to evaluate various cognitive functions. It is important to note that cognitive decline is not a strictly binary condition; rather, it is a progressive process that is more accurately represented on a continuous scale.

To further evaluate our method in estimating the MMSE score, we conducted a regression task. The results, obtained through 5-fold cross-validation, yielded a RMSE of 4.77. Among the works referenced for comparison in the classification task, only the study in~\cite{rohanian21_interspeech} reports results for the regression task, achieving an RMSE of 4.26. However, the cross-validation protocol used in that study is not explicitly stated, limiting direct comparability. Additionally, the baseline model proposed in ADReSSo dataset was evaluated using LOSO cross-validation, resulting in a best RMSE of 5.52. Under the same LOSO evaluation protocol, our method achieves an RMSE of 5.28, thereby outperforming the baseline approach.

\subsection{Test on external dataset}

In addition to the main evaluation presented over the ADReSSo dataset, we further examined the generalizability of our approach using data derived from the DementiaBank corpora. While the Pitt Corpus~\cite{lanzi2023dementiabank} remains the most widely used benchmark in speech-based dementia research, several of its original recordings contain residual tape noise and other artefacts that may distort acoustic features. To address these limitations, we employed the Alzheimer’s Dementia Recognition through Spontaneous Speech (ADReSS) dataset~\cite{luz20_interspeech}, a carefully curated, acoustically pre-processed, and demographically balanced subset of the Pitt  Corpus, and conducted a 5-fold cross-validation experiment. Although ADReSS includes manual transcripts, we regenerated them using the Whisper ASR model to obtain the word-level timestamps required by our alignment module.

CogniAlign achieved an accuracy of 85.06\% on this dataset across five-fold cross-validation, demonstrating the method’s generalizability. The current state-of-the-art performance on this dataset is reported in~\cite{ilias2022multimodal}, where the authors achieved 90\% accuracy. Notably, their approach used the original manual transcriptions, which typically led to improved results, as also shown by the ADReSSo baseline. As with the regression task, no ablation studies or hyperparameter tuning were performed for this evaluation.

\subsection{Ablation Study}\label{subsec:ablation}

An extensive ablation study has been performed to identify the most promising and relevant configurations among the proposed approaches. This subsection presents a detailed analysis of the evaluated embeddings, fusion strategies, and key hyperparameters. All experiments presented in this section were conducted using a 5-fold cross-validation setup.

\begin{table}[htpb]
\centering
\begin{tabular}{lcccc}
\toprule
Model & Accuracy & F1 Score & Recall & Precision \\\midrule
BERT              & 84.39 & 83.92 & 83.82 & 84.49 \\
DistilBERT        & \textbf{84.97} & \textbf{84.75} & \textbf{85.07} & \textbf{84.88} \\
RoBERTa           & 79.55 & 78.83 & 79.58 & 81.13 \\
Mistral           & 81.28 & 80.98 & 81.38 & 81.17 \\
Qwen              & 83.12 & 83.00 & 83.67 & 83.30 \\
Stella            & 84.33 & 83.69 & 84.05 & 84.64 \\\bottomrule
\end{tabular}
\caption{Performance of text-based models on five-fold cross-validation using the ADReSSo dataset. Bold values denote the best performance.}
\label{tab:text}
\end{table}

\vspace{1em}
\noindent\textbf{Text Modality}\\
To identify the most effective textual encoder, we evaluated a diverse set of pretrained language models, including BERT, DistilBERT, RoBERTa, Mistral (\emph{Mistral-7B-v0.1}), Qwen (\emph{Qwen2.5-7B}), and Stella (\emph{stella\_en\_1.5B\_v5}). These models were selected to represent a range of architectures and sizes.

The results, summarized in Table~\ref{tab:text}, indicate that \emph{DistilBERT} achieved the highest performance among the evaluated models. Interestingly, larger language models such as Mistral, Qwen, and Stella did not outperform the more compact architectures like BERT and DistilBERT. This outcome may be attributed to the relatively small size of the ADReSSo dataset and the presence of atypical speech patterns commonly observed in individuals with Alzheimer’s disease. Moreover, these larger models are primarily trained for generative tasks, whereas models like DistilBERT are optimized for discriminative tasks, such as classification, during pretraining. This distinction may better align with our objectives in cognitive assessment.

\begin{table}[htpb]
\centering
\begin{tabular}{lcccc}
\toprule
Model & Accuracy & F1 Score & Recall & Precision \\\midrule
eGeMAPS         & 65.69 & 59.89 & 64.62 & 63.96 \\
Mel Spec.       & 60.23 & 52.15 & 58.74 & 61.09 \\
Wav2Vec2      & \textbf{79.52} & \textbf{78.76} & \textbf{78.69} & \textbf{80.01} \\\bottomrule
\end{tabular}
\caption{Performance of audio-based models on five-fold cross-validation using the ADReSSo dataset. Bold values indicate the best performance.}
\label{tab:audio}
\end{table}

\noindent\\\textbf{Audio Modality}\\
To evaluate the performance of different audio representations, we tested three feature extraction strategies: eGeMAPS acoustic features~\cite{7160715} extracted using OpenSMILE tooklkit~\cite{10.1145/1873951.1874246}, Mel spectrograms~\cite{satt2017efficient,zeng2019spectrogram}, and Wav2Vec2 embeddings. The results are presented in Table~\ref{tab:audio}.

Among these, the word-level aligned Wav2Vec2 model achieved the best performance of the baseline variants. While Mel spectrograms and eGeMAPS embeddings have also been aligned at the word level, they showed inferior performance compared to Wav2Vec2. Nonetheless, their inclusion demonstrates the generalizability of the proposed alignment strategy across heterogeneous acoustic representations. Specifically, Mel spectrograms were computed using 20ms non-overlapping windows to match the temporal resolution of the Wav2Vec2 model. Meanwhile, eGeMAPS features were extracted using 100ms windows, the minimum supported by the OpenSMILE toolkit.

\begin{table}[htpb]
\centering
\begin{tabular}{lcccc}
\toprule
Model & Accuracy & F1 Score & Recall & Precision \\\midrule
DistilBERT & 84.97 & 84.75 & 85.07 & 84.88 \\
DistilBERT P & \textbf{86.77} & \textbf{86.59} & \textbf{87.07} & \textbf{86.83} \\
\hdashline
Wav2Vec2 & 79.52 & 78.76 & 78.69 & 80.01 \\
Wav2Vec2 P & \textbf{80.12} & \textbf{79.46} & \textbf{79.88} & \textbf{81.40} \\\bottomrule
\end{tabular}
\caption{Performance of unimodal models with and without prosodic information on five-fold cross-validation using the ADReSSo dataset. Models with names ending in ``P'' include prosodic features. Bold values indicate the best overall performance within each modality.}
\label{tab:prosodic}
\end{table}

\vspace{1em}
\noindent\textbf{Prosodic information}\\
Building upon the previously identified best-performing unimodal models, DistilBERT for the textual modality and Wav2Vec2 for the audio modality, we investigate the impact of incorporating prosodic information, specifically speech pauses, into these architectures. In what follows, we refer to models augmented with prosodic features using the suffix ``P'' (e.g., DistilBERT P), and the results are presented in Table~\ref{tab:prosodic}.

In the enhanced textual model, DistilBERT P, pause tokens were inserted into the input sequence to reflect temporal gaps in speech. This augmentation led to a consistent improvement in performance, emphasizing that prosodic cues, such as hesitation and silence, can carry meaningful diagnostic information even when no linguistic content is present.

Similarly, we developed an enhanced audio model, Wav2Vec2 P, by enriching the base Wav2Vec2 architecture with prosodic features. This modification resulted in the highest performance among our unimodal audio model, further confirming that periods of silence, often overlooked, can provide valuable insights into cognitive decline.

The consistent performance improvements observed across both modalities underscore the critical role of prosodic information in Alzheimer's disease detection. Notably, these enhancements enable the analysis of patient behavior not only during active speech but also in moments of silence or hesitation, phenomena that are often symptomatic yet underexplored.

These two enhanced models, DistilBERT P and Wav2Vec2 P, are the versions used for comparison with existing state-of-the-art unimodal approaches, as previously reported in Table~\ref{tab:sota}.

\begin{table}[htpb]
\centering
\begin{tabular}{lcccc}
\toprule
Fusion & Accuracy & F1 Score & Recall & Precision \\\midrule
Concat  & 87.36 & 87.05 & 87.08 & 87.31 \\
Mean    & 86.76 & 86.54 & 86.74 & 86.53 \\
Prod     & 87.34 & 87.14 & 87.34 & 87.42 \\
Sum     & 86.76 & 86.54 & 86.74 & 86.53 \\
SA   & 86.17 & 85.99 & 86.48 & 86.30 \\
CA   & 88.54 & 88.23 & 88.13 & 88.71 \\
GCA  & \textbf{90.36} & \textbf{90.11} & \textbf{90.15} & \textbf{90.76} \\
BCA  & 87.93 & 87.71 & 87.69 & 87.93 \\
GBCA & 89.14 & 88.90 & 88.84 & 89.30 \\\bottomrule
\end{tabular}
\caption{
Performance of proposed fusion strategies applied to the multimodal models. \emph{Concat} denotes feature concatenation. \emph{Mean}, \emph{Sum}, and \emph{Prod} (element-wise product) represent arithmetic fusion strategies. For simplicity, Self-Attention (\emph{SA}), Cross-Attention (\emph{CA}), Gated Cross-Attention (\emph{GCA}), Bidirectional Cross-Attention (\emph{BCA}), and its gated variant (\emph{GBCAG}). Bold values indicate the best overall performance.}

\label{tab:fusion_strategies}
\end{table}

\noindent\\\textbf{Multimodal Fusion Strategy}\\
Following the identification of the best-performing unimodal models, DistilBERT P for text and Wav2Vec2 P for audio, we evaluated several multimodal fusion strategies to determine the most effective mechanism for combining aligned word-level embeddings. Each fusion method leverages the word-level alignment process described earlier, ensuring synchronized token sequences from both modalities.

\begin{figure*}[htbp]
    \centering
    \begin{subfigure}[b]{0.15\textwidth}
        \includegraphics[width=\linewidth]{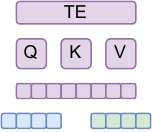}
        \caption*{(a)}
    \end{subfigure}
    \hspace{0.01\textwidth}
    \begin{subfigure}[b]{0.15\textwidth}
        \includegraphics[width=\linewidth]{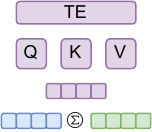}
        \caption*{(b)}
    \end{subfigure}
    \hspace{0.01\textwidth}
    \begin{subfigure}[b]{0.15\textwidth}
        \includegraphics[width=\linewidth]{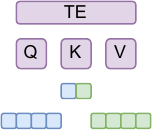}
        \caption*{(c)}
    \end{subfigure}
    \hspace{0.01\textwidth}
    \begin{subfigure}[b]{0.15\textwidth}
        \includegraphics[width=\linewidth]{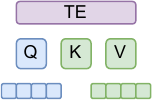}
        \caption*{(d)}
    \end{subfigure}
    \hspace{0.01\textwidth}
    \begin{subfigure}[b]{0.3\textwidth}
        \includegraphics[width=\linewidth]{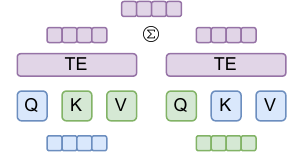}
        \caption*{(e)}
    \end{subfigure}
    \caption{
Transformer-based fusion strategies explored in this work: 
(a) Concatenation, 
(b) Element-wise Fusion (e.g., Sum, Product, or Mean), 
(c) Self-Attention Fusion, 
(d) Cross-Attention Fusion (including gated variant), 
and (e) Bidirectional Cross-Attention Fusion (including gated variant). Blue blocks represent one input modality, green blocks represent the other modality, and purple blocks correspond to fused multimodal representations. 
Better viewed in color.
}
\label{fig:fusions}
\end{figure*}

We evaluated five principal fusion strategies (see Figure~\ref{fig:fusions}):
\begin{itemize}
    \item \textbf{Concatenation:} The embeddings from both modalities are concatenated and passed through a shared Transformer Encoder. This strategy preserves the full representation from each modality but lacks explicit modeling of inter-modality interactions~\cite{guo2020graphcodebert,shi2022learningaudiovisualspeechrepresentation}.
    
    \item \textbf{Element-wise Fusion:} Arithmetic operations, such as sum, mean, or element-wise product, combine token embeddings directly. This method requires that both modalities have sequences of equal length and dimensionality~\cite{9156959,10377257}.
    
    \item \textbf{Self-Attention Fusion:} Each modality is first compressed into a single token via mean pooling. These two tokens (one for each modality) are then processed by a Transformer encoder with self-attention, which captures interactions between the summary embeddings~\cite{chumachenko2022self,10.1007/978-3-031-43895-0_15}.
    
    \item \textbf{Cross-Attention Fusion:} Token sequences from one modality attend to those of the other through a cross-attention Transformer Encoder. In this fusion, Audio modality serves as the query, while textual embeddings act as key/value inputs to a Transformer Encoder. This configuration facilitates fine-grained, temporally aligned interactions between modalities~\cite{murahari2020large,lin2020interbert}. A gated variant is also evaluated, where a learnable sigmoid gate modulates the contribution of the attended features before combining them with the original audio representation~\cite{electronics13112046,lee2022leaky}. A detailed description of the gating mechanism was presented in Section~\ref{sec:method}.
    
    \item \textbf{Bidirectional Cross-Attention Fusion:} Two parallel Transformer Encoders are employed: one where audio is the query and text the key/value, and another with the roles reversed. The resulting representations from both attention directions are averaged via a mean pooling to form the final fused embedding~\cite{yang-wang-2024-figclip,wang2022domain}. As in the unidirectional case, a gated variant is implemented for each branch, dynamically controlling the integration of cross-modal features.
\end{itemize}

Table~\ref{tab:fusion_strategies} presents the performance of each approach. Overall, the Gated Cross-Attention Fusion model yielded the highest accuracy (90.36\%), clearly outperforming both unimodal baselines and all other fusion strategies.

Conventional approaches such as concatenation improved slightly over unimodal models, highlighting the benefit of combining modalities. However, element-wise operations generally performed worse, likely due to their inability to preserve the richness of each modality. Self-attention fusion also underperformed, which we attribute to the loss of contextual detail during the pooling step that reduces each modality to a single vector.

Cross-attention strategies yielded the best results. Their ability to model token-level dependencies between temporally aligned inputs proved critical for capturing multimodal nuances. The gated variant further enhanced this by enabling dynamic control over the contribution of attended features, filtering out noise, and reinforcing relevant interactions. These results mark a substantial improvement over unimodal baselines, underscoring the benefits of this multimodal fusion in capturing complementary cues from both speech content and acoustic patterns.

The addition of the gating mechanism not only enhances the performance of the cross-attention model but also yields improvements in its bidirectional variant. To ensure that the observed performance gains from the gated fusion are not attributable to sampling noise, we conducted a paired t-test~\cite{kim2015t} on 50 accuracy pairs derived from five random seeds combined with ten-fold stratified cross-validation.

The test revealed a mean accuracy The test revealed a mean accuracy improvement of 1.36 percentage points for the gated cross-attention model over the standard cross-attention version, which was statistically significant (p < 0.05). This metric allows us to conclude that the gating mechanism provides a statistically significant advantage over the standard cross-attention approach~\cite{de2013using}.

The Bidirectional Cross-Attention (BCA) model also performed well but did not surpass the single-direction Gated Cross-Attention (GCA) approach. This lower result may be due to the asymmetric effectiveness of each modality when used as a query or key/value. Specifically, prior results show that textual features are more informative than audio. Therefore, using text as a key/value allows the model to attend to richer semantic content guided by the prosodic structure of the audio.

In contrast, the second attention direction in the bidirectional setup, where audio serves as key/value and text as the query, may offer less meaningful guidance, limiting the benefit of this path. Averaging both attention directions could, therefore, reduce the overall effectiveness of the model.

\noindent\\\textbf{Cross-Attention Directional Query}\\
To better understand the effect of attention directionality, we conducted an ablation comparing the effect of query directionality in the Gated Cross-Attention architecture. As shown in Table~\ref{tab:cross}, using audio as the query and text as key/value achieves the best results.

This finding aligns with theoretical expectations: in the cross-attention mechanism, the key/value modality is the primary source of contextual information, while the query guides attention. Since the textual encoder consistently outperforms the audio encoder in unimodal tasks, preserving textual features as the key/value source enables more meaningful and stable attention. The audio modality, though less informative in isolation, provides essential prosodic and paralinguistic cues that can guide the model's focus during fusion.

\begin{table}[htpb]
\centering
\begin{tabular}{lccccc}
\toprule
Query & Accuracy & F1 Score & Recall & Precision \\\midrule
DistilBERT P & 84.88 & 84.75 & 85.16 & 85.33 \\
Wav2Vec2 P & \textbf{90.36} & \textbf{90.11} & \textbf{90.15} & \textbf{90.76} \\\bottomrule
\end{tabular}
\caption{
Comparison of the Query modality for the Gated Cross-Attention Multimodal Fusion. Each row shows the results when the specified model is used as the query, while the other modality is used as key and value. Bold values represent the best overall performance.}
\label{tab:cross}
\end{table}

Conversely, using text as the query and audio as the key/value yields inferior results. This direction limits the model's ability to extract the rich semantic structure embedded in the textual embeddings, as it must attend to the comparatively noisier and lower-level audio space. This finding also explains the slight drop in performance of the Bidirectional Cross-Attention model, where averaging includes the less effective attention direction. By introducing both directions symmetrically, the bidirectional model potentially suppresses the more informative text-guided attention path.

\begin{table}[htpb]
\centering
\begin{tabular}{lcccc}
\toprule
Pooling & Accuracy & F1 Score & Recall & Precision \\\midrule
Attn     & 86.17 & 85.97 & 86.26 & 85.96 \\
CLS      & 87.38 & 87.21 & 87.90 & 88.14 \\
GatedAttn & 89.77 & 89.52 & 89.67 & 90.01 \\
Mean     & \textbf{90.36} & \textbf{90.11} & \textbf{90.15} & \textbf{90.76} \\\bottomrule
\end{tabular}
\caption{
Comparison of pooling strategies for the Gated Cross-Attention Multimodal Fusion. \emph{GatedAttn} denotes Gated Attention Pooling and \emph{Attn} represents Attention Pooling.}
\label{tab:pooling_strategies}
\end{table}

\noindent\\\textbf{Pooling Strategy}\\
From the best-performing model configuration, we conducted an ablation study on the Transformer Encoder architecture to assess the impact of different token aggregation strategies for final classification. Specifically, we compared several pooling methods used to generate the input to the MLP classifier. These included the use of the \texttt{[CLS]} token, originally introduced in BERT and now widely adopted across Transformer-based architectures~\cite{dosovitskiy2020image,chen2021crossvit}, standard attention-based pooling, its gated variant, and mean pooling across all token embeddings. As shown in Table~\ref{tab:pooling_strategies}, mean pooling consistently achieved the highest performance.

\begin{table}[htpb]
\centering
\begin{tabular}{lcccc}
\toprule
N Layers & Accuracy & F1 Score & Recall & Precision \\\midrule
1 & \textbf{90.36} & \textbf{90.11} & \textbf{90.15} & \textbf{90.76} \\
2 & 88.56 & 88.21 & 88.23 & 89.43 \\
3 & 88.57 & 88.05 & 87.89 & 89.30 \\
4 & 89.77 & 89.43 & 89.38 & 90.29 \\\bottomrule
\end{tabular}
\caption{
Effect of varying the number of Transformer Encoder layers in the Gated Cross-Attention Multimodal Fusion.}
\label{tab:nlayers_ablation}
\end{table}

\noindent\\\textbf{Number of Transformer Encoder Layers}\\
The effect of varying the number of Transformer Encoder layers has also been examined, with results shown in Table~\ref{tab:nlayers_ablation}. These findings indicate that a single Transformer Encoder layer yields the best performance. This outcome is likely because each modality is already processed by a specialized unimodal model, which extracts rich, high-level features. Consequently, adding more Transformer layers increases the complexity of the model leading to decreased performance on this dataset.

\section{Explainability}\label{sec:explainable}

In healthcare-related tasks, it is essential to understand how models make decisions, particularly which features or patterns they rely on when predicting Alzheimer’s disease or assessing cognitive health, to ensure explainability~\cite{tjoa2020survey,yang2022unbox}. This information can also be valuable for healthcare professionals, offering insights into the model’s reasoning and potentially supporting clinical interpretation. Accordingly, in this section, we investigate the explainability of our approach by conducting a statistical corpus analysis of the ADReSSo dataset and examining the attention mechanisms within the models.

\subsection{Prosodic information}

In this work, we incorporate prosodic cues into the original textual transcriptions by adding pauses. As previously noted, one of the common symptoms of Alzheimer’s disease is word-finding difficulty. This often manifests as hesitations in speech, where the speaker pauses to search for the appropriate word. These pauses are typically represented in transcriptions using commas, periods, or ellipses, depending on their duration~\cite{PISTONO2019133}. To better capture this behavior, we categorized pauses in the patients’ speech into three types based on their duration. The inclusion of these pause types has led to improved model performance, as demonstrated in the ablation study and supported by findings in related research~\cite{yuan20_interspeech,zhu21e_interspeech,rohanian21_interspeech}. Additionally, a corpus-level analysis of these pauses was conducted, with results presented in Table~\ref{tab:prosodicstats}.

\begin{table}[htpb]
    \centering
    \begin{tabular}{lcccc}
    \toprule
    \multirow{2}{*}{Pause} & \multicolumn{2}{c}{Original} & \multicolumn{2}{c}{Additional}
    \\
    \cline{2-5}
    & CH & AD & CH & AD \\
    \midrule
    ``,'' & \textbf{5.49} & 5.17 & \textbf{2.27} & 1.85 \\
    ``.'' & 11.85 & \textbf{13.94} & 2.67 & \textbf{2.85} \\
    ``...'' & 0.29 & \textbf{1.06} & 3.15 & \textbf{5.85} \\
    \bottomrule
    \end{tabular}
    \caption{Average (mean) number of occurrences per transcription for different pause types. Values are reported separately for cognitively healthy (CH) and Alzheimer's disease (AD) participants, in both the original transcriptions and the additional prosodic annotations. Bold values indicate the higher mean within each row.}
    \label{tab:prosodicstats}
\end{table}

The results reveal a tendency for cognitively healthy individuals to produce shorter pauses, as observed in both the original and the additional transcriptions. These shorter pauses, typically represented by commas, are embedded within sentences and do not necessarily indicate hesitation, as they reflect only brief interruptions in speech.

In contrast, longer pauses are more frequently associated with individuals diagnosed with Alzheimer’s disease. This trend becomes increasingly evident with longer pauses, particularly those marked by ellipses. A similar pattern was reported in~\cite{PASTORIZADOMINGUEZ2022107}, where healthy participants exhibited a higher frequency of short pauses. On the other hand, individuals with Alzheimer’s showed a greater tendency toward longer pauses.

These findings are further supported by differences in the average length of audio recordings. In this dataset, the mean duration of audio samples for participants with Alzheimer’s disease is 87.61 seconds, compared to 68.76 seconds for cognitively healthy individuals. This contrast is even more pronounced when considering the number of words per transcription: healthy individuals average 97.16 words per sample, while those with Alzheimer’s average 118.96 words. This suggests that despite producing longer recordings, individuals with Alzheimer’s may require more time and pauses to produce fewer words.

\subsection{Model Explainability}

\begin{figure*}
    \centering
    \begin{subfigure}[b]{0.8\textwidth}
        \includegraphics[width=\linewidth]{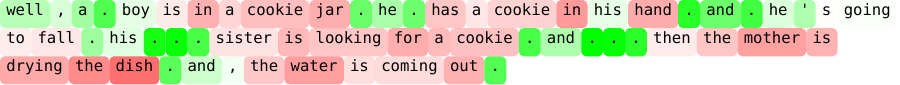}
        \caption*{(a) Influence for the prediction of Alzheimer's disease.}
    \end{subfigure}    
    \vspace{0.2cm}
    
    \begin{subfigure}[b]{0.8\textwidth}
        \includegraphics[width=\linewidth]{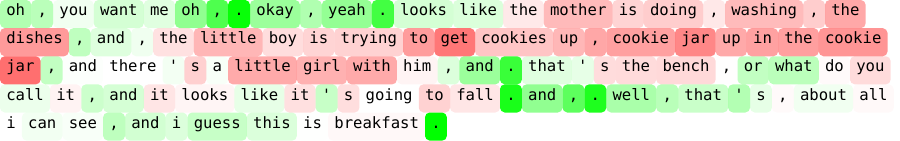}
        \caption*{(b) Influence for the prediction of Alzheimer's disease.}
    \end{subfigure}
    \vspace{0.2cm}

    \begin{subfigure}[b]{0.8\textwidth}
        \includegraphics[width=\linewidth]{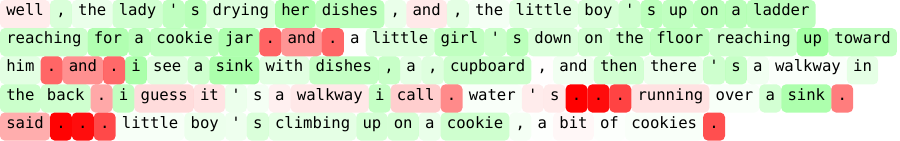}
        \caption*{(c) Influence for the prediction of cognitive healthy.}
    \end{subfigure}
    \vspace{0.2cm}

    \begin{subfigure}[b]{0.8\textwidth}
        \includegraphics[width=\linewidth]{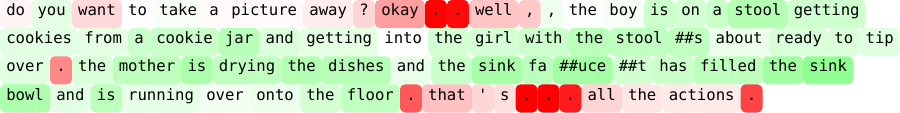}
        \caption*{(d) Influence for the prediction of Alzheimer's disease.}
    \end{subfigure}
    \vspace{0.2cm}
    
    \caption{Integrated Gradients attribution visualization from CogniAlign Model. Tokens contributing positively to the predicted class are shown in green, while those reducing its likelihood are in red. Color intensity corresponds to the relative importance of each token.}
    \label{fig:igvis}
\end{figure*}

To better understand the contribution of each input token to the model’s final prediction, we applied Integrated Gradients (IG)~\cite{sundararajan2017axiomatic}, a gradient-based attribution method. This approach quantifies the influence of individual input features by integrating the gradients of the model’s output along a linear path from a baseline input (typically a zero vector) to the actual input. In our setup, IG has been employed to compute token-level attribution scores for each prediction, representing relative importance. The results of this analysis are presented in Figure~\ref{fig:igvis}.

Based on the results obtained from the evaluation of the full CogniAlign architecture with prosodic enhancement, we observe a strong influence of prosodic cues on the model’s predictions. Specifically, these cues increase the model’s propensity to predict Alzheimer’s disease while simultaneously reducing its propensity to predict cognitive health. Such markers are commonly associated with individuals experiencing word-finding difficulties or requiring additional time to organize their thoughts, thereby offering both interpretability and clinical relevance.

Regarding lexical features, certain discourse markers such as ``okay'' and ``well'' also exhibit a positive influence on the prediction of Alzheimer’s disease. This observation is consistent with our earlier analysis of prosodic cue distributions in the dataset. Similar patterns have been reported in prior studies in the field~\cite{ORTIZPEREZ2023126413, ilias2022explainable}, further validating our findings. These insights not only support the model’s behavior but also align with established clinical observations~\cite{almor1999alzheimer}, which report that individuals with Alzheimer’s disease frequently exhibit longer pauses, hesitations, and disrupted discourse structure due to lexical retrieval difficulties. The model’s reliance on prosodic cues such as extended silences and discourse markers like ``okay'' or ``well'' reflects these clinical symptoms, thereby enhancing both its interpretability and practical utility in a healthcare context.
\section{Limitations}\label{sec:limitations}

While our methodology, along with other state-of-the-art approaches, demonstrates significant and relevant results, a clear limitation remains in this research field: the lack of large, diverse datasets suitable for this task. The primary reason for this limitation is the sensitivity of the data, particularly audio recordings of individuals with cognitive impairments, which significantly restricts access and sharing.

Creating such datasets is a complex and time-consuming process. In this domain, the challenges are even greater, as data collection must involve healthcare professionals to design and oversee the recording protocols. Moreover, navigating the bureaucratic procedures and obtaining the necessary permissions to handle and share sensitive data, even for academic purposes, presents an additional barrier.

This limitation in dataset size and diversity affects not only the performance of current studies but also the feasibility of deploying these models in real-world scenarios. Small and homogeneous datasets increase the risk of overfitting, where models perform well on evaluation metrics but fail to generalize to broader, real-world populations. Furthermore, many of the existing datasets are collected from narrow demographic groups, often limited to a single language or culture. Without sufficient representation of diverse cultures and languages, the deployment of these algorithms beyond the original context becomes unfeasible.

Therefore, there is a critical need for the development of large-scale, multimodal, and multicultural datasets. Such resources are essential to ensure robust, fair, and generalizable training of deep learning models in this sensitive and impactful area.
\section{Conclusions}\label{sec:conclusions}

In this work, we introduced CogniAlign, a multimodal model for Alzheimer's disease detection that leverages a word-level alignment strategy to bridge audio and textual modalities. The model achieves semantically consistent and temporally grounded fusion by synchronizing audio embeddings with the corresponding textual tokens through precise temporal alignment. This alignment facilitates the use of advanced attention-based mechanisms, allowing the model to capture fine-grained cross-modal interactions. In particular, CogniAlign incorporates a Gated Cross-Attention fusion strategy, which has proven especially effective in enhancing integration between modalities.

Beyond alignment, the model also integrates prosodic information by detecting and encoding interword pauses, which are inserted as explicit tokens in both the textual and audio streams. These prosodic cues offer additional insight into speech hesitations and rhythm, improving performance across both modalities. Our experimental results, conducted on the ADReSSo dataset, demonstrate the superiority of the proposed architecture over unimodal baselines and state-of-the-art methods. CogniAlign achieved an accuracy of 90.36\%, outperforming existing approaches and establishing a new benchmark for multimodal Alzheimer's detection.

The ablation studies presented throughout this work further validate the design choices of the architecture. These analyses confirm the effectiveness of cross-attention mechanisms over simpler alternatives, such as concatenation or element-wise operations. The gated variant of cross-attention yielded the most substantial gains, emphasizing the value of selectively integrating information from each modality. Additionally, our investigation into attention directionality highlights the importance of using textual features as the key/value source, given their stronger unimodal performance and semantic richness.

Several important insights emerge from our findings. First, temporal alignment at the word level is a critical enabler for coherent multimodal fusion, especially when combined with attention-based strategies. Second, prosodic features, particularly pause patterns, serve as a valuable signal in both modalities, enhancing the model’s capacity to detect cognitive decline. Finally, our results reaffirm the dominant role of textual information in this task, both in unimodal models and in guiding attention in multimodal fusion.

In future work, we plan to extend this study by applying the proposed methodologies to other disease domains to evaluate the generalizability of our approach. Based on the current results, the inclusion and analysis of additional modalities, such as video, will also be explored. Finally, given the existing alignment between audio and textual modalities, we intend to investigate a weighted fusion of corresponding tokens to better understand the relative influence and impact of different spoken and written elements.

\section*{Acknowledgment}
This work was supported by the Spanish State Research Agency (AEI) under grant: GEMELIA PID2024-161711OB-I00; and ERDF/EU. Additionally, it received support from the CIAICO/2022/132 Consolidated group project ``AI4Health'',funded by the Valencian government and International Center for Aging Research ICAR funded project ``IASISTEM''. This work is also a part of the ENIA Chair of Artificial Intelligence from the University of Alicante (TSI-100927-2023-6) funded by the Recovery, Transformation and Resilience Plan from the European Union Next Generation through the Ministry for Digital Transformation and the Civil Service. This work has also been supported by two Spanish national and a regional grants for PhD studies, FPU21/00414, FPU23/00532 and CIACIF/2022/175.

\bibliographystyle{ieeetr}
\bibliography{bibliography}

\end{document}